# "Squeaky Wheel" Optimization

**David E. Joslin**                                                    DAVID_JOSLIN@I2.COM
*i2 Technologies*
*909 E. Las Colinas Blvd.*
*Irving, TX 75039*

**David P. Clements**                                       CLEMENTS@CIRL.UOREGON.EDU
*Computational Intelligence Research Laboratory*
*1269 University of Oregon*
*Eugene, OR 97403-1269*

## Abstract

We describe a general approach to optimization which we term "Squeaky Wheel" Optimization (SWO). In SWO, a greedy algorithm is used to construct a solution which is then analyzed to find the trouble spots, i.e., those elements, that, if improved, are likely to improve the objective function score. The results of the analysis are used to generate new priorities that determine the order in which the greedy algorithm constructs the next solution. This Construct/Analyze/Prioritize cycle continues until some limit is reached, or an acceptable solution is found.

SWO can be viewed as operating on two search spaces: solutions and prioritizations. Successive solutions are only indirectly related, via the re-prioritization that results from analyzing the prior solution. Similarly, successive prioritizations are generated by constructing and analyzing solutions. This "coupled search" has some interesting properties, which we discuss.

We report encouraging experimental results on two domains, scheduling problems that arise in fiber-optic cable manufacturing, and graph coloring problems. The fact that these domains are very different supports our claim that SWO is a general technique for optimization.

## 1. Overview

We describe a general approach to optimization which we term "Squeaky Wheel" Optimization (SWO) (Joslin & Clements, 1998). The core of SWO is a Construct/Analyze/Prioritize cycle, illustrated in Figure 1. A solution is constructed by a greedy algorithm, making decisions in an order determined by priorities assigned to the elements of the problem. That solution is then analyzed to find the elements of the problem that are "trouble makers." The priorities of the trouble makers are then increased, causing the greedy constructor to deal with them sooner on the next iteration. This cycle repeats until a termination condition occurs.

On each iteration, the analyzer determines which elements of the problem are causing the most trouble in the current solution, and the prioritizer ensures that the constructor gives more attention to those elements on the next iteration. ("The squeaky wheel gets the grease.") The construction, analysis and prioritization are all in terms of the elements that define a problem domain. In a scheduling domain, for example, those elements might be tasks. In graph coloring, those elements might be the nodes to be colored.

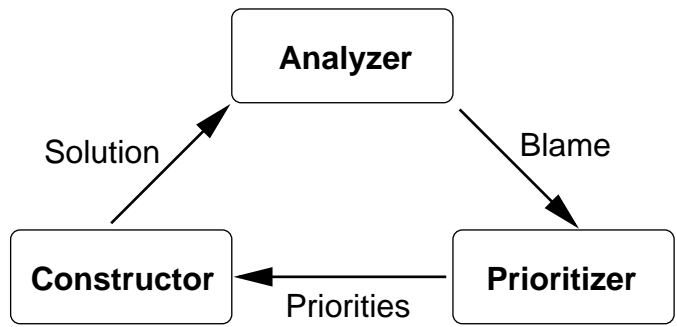

Figure 1: The Construct/Analyze/Prioritize cycle

The three main components of SWO are:

**Constructor.** Given a sequence of problem elements, the constructor generates a solution using a greedy algorithm, with no backtracking. The sequence determines the order in which decisions are made, and can be thought of as a "strategy" or "recipe" for constructing a new solution. (This "solution" may violate hard constraints.)

**Analyzer.** The analyzer assigns a numeric "blame" factor to the problem elements that contribute to flaws in the current solution. For example, if minimizing lateness in a scheduling problem is one of the objectives, then blame would be assigned to late tasks.

A key principle behind SWO is that solutions can reveal problem structure. By analyzing a solution, we can often identify elements of that solution that work well, and elements that work poorly. A resource that is used at full capacity, for example, may represent a bottleneck. This information about problem structure is local, in that it may only apply to the part of the search space currently under examination, but may be useful in determining where the search should go next.

**Prioritizer.** The prioritizer uses the blame factors assigned by the analyzer to modify the previous sequence of problem elements. Elements that received blame are moved toward the front of the sequence. The higher the blame, the further the element is moved.

The priority sequence plays a key role in SWO. As a difficult problem element moves forward in the sequence it is handled sooner by the constructor. It also tends to be handled better, thus decreasing its blame factor. Difficult elements rise rapidly to a place in the sequence where they are handled well. Once there, the blame assigned to them drops, causing them to slowly sink in the sequence as other parts of the problem that are not handled as well are given increased priority. Eventually, difficult elements sink back to the point where they are no longer handled well, causing them to receive higher blame and to move forward in the sequence again. Elements that are always easy to handle sink to the end of the sequence and stay there.

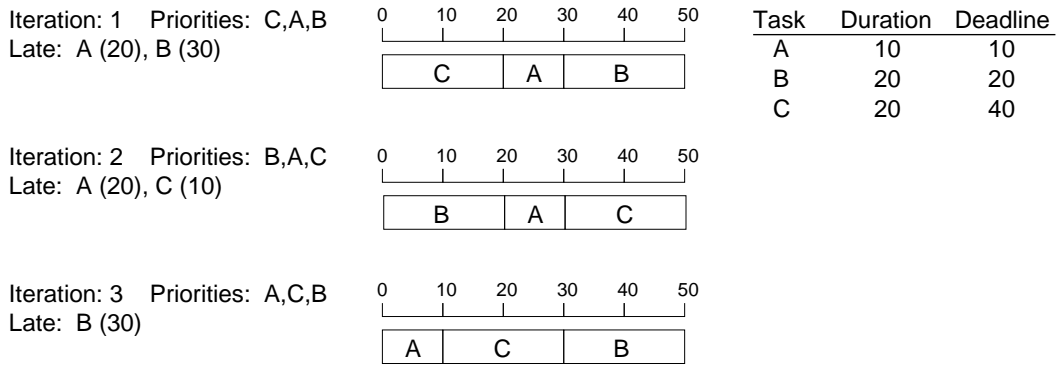

Figure 2: Simple example

To illustrate the SWO cycle, consider a simplified scheduling example. Suppose we have a single production line, and three tasks to schedule, $A$, $B$ and $C$. Only one task can be performed at a time. Execution starts at $t = 0$. The duration and deadline for each task is shown in Figure 2. The objective is to minimize the number of late tasks. An optimal solution will have one late task.

Suppose our initial priority sequence is $\langle C, A, B \rangle$, and the constructor schedules tasks in order, at the earliest possible time. The resulting schedule has two late tasks ($B$ and $A$). Suppose that our analyzer assigns one point of "blame" to each late task, for each unit of time it is late. In this case, $A$, $B$, and $C$ receive 20, 30 and 0 units of blame, respectively. Figure 2 shows the prioritization, the schedule that the constructor builds from that prioritization, and the late tasks with the blame assigned to each.

For the next cycle, the prioritizer must take the previous priority sequence, and the blame assigned by the analyzer, and generate a new priority sequence. A simple prioritizer might just sort the tasks by their numeric blame in descending order, resulting in the new priority sequence $\langle B, A, C \rangle$.

After the second cycle, tasks $A$ and $C$ are late, scoring 20 and 10 points of blame, respectively. The new priority sequence is then $\langle A, C, B \rangle$.

The third solution, constructed from this priority sequence, has only one late task, $B$, which receives 30 points of blame. At this point we have an optimal solution. If we continue running SWO, however, as we might expect to do since we typically do not know when we have reached optimality, SWO will attempt to fix what was wrong with the current solution. Here, since task $B$ was late, its priority would be increased, and the resulting solution would fix that problem at the expense of others. (We would also enter a short cycle, alternating between the last two schedules. We address this by introducing some randomization in the prioritizer.)

Although this example is highly simplified, and there would clearly be better and more sophisticated ways to implement each of the three modules, Figure 3 shows that the behavior illustrated by the simple example is reflected in a real domain. The figure shows the changing position in the priority sequence of three tasks in the scheduling domain that is described in detail in the following section. One task ("Job 24") starts out with a high priority, and remains at a relatively high priority level. We can see that when the task is scheduled effectively, and therefore receives little or no blame, its priority tends to drop, but it does not have to drop very far before it ceases to be scheduled well, acquires a significant level of blame, and moves quickly back to a higher priority.

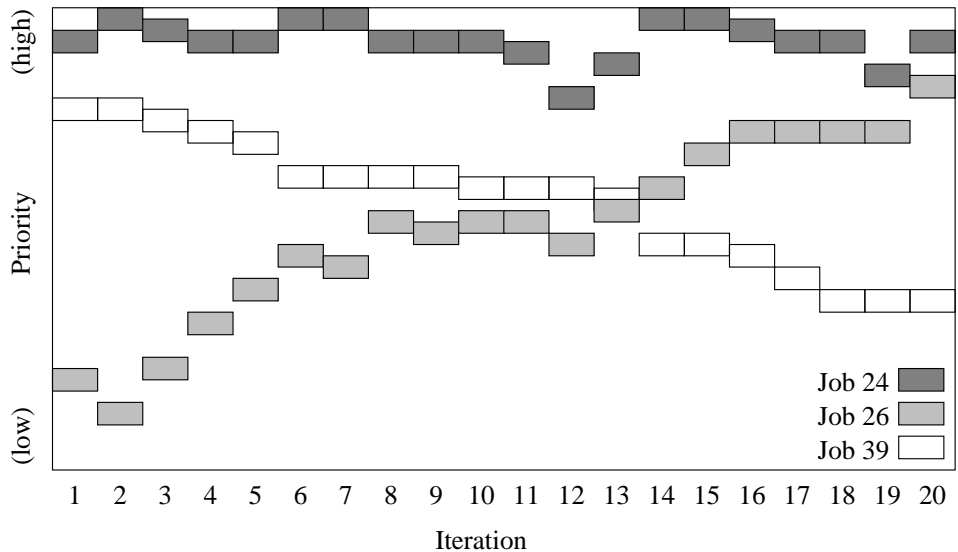

Figure 3: Examples of priority changes over time

The other two tasks shown in the figure behave quite differently. One task ("Job 39") starts out with a relatively high priority, but this task is "easy" to schedule, with little blame, even when it is scheduled late in the sequence. Over successive iterations, the priority of such a task will tend to decrease steadily. The other task illustrated here ("Job 26") does just the opposite, starting at a low priority and moving fairly steadily toward a high priority.

The following section discusses the characteristics of SWO that make it an effective technique for optimization. We then discuss implementations of SWO for scheduling and for graph coloring problems. The final sections discuss related work, describe directions for future research, and summarize our findings.

## 2. Key ideas

As the experimental results below show, SWO is a general approach to optimization. In this section, we explore a few insights into what makes SWO effective.

It is useful to think of SWO as searching two coupled spaces, as illustrated in Figure 4. One search space is the familiar solution space, and the other is *priority space*. Moves in the solution space are made indirectly, via the re-prioritization that results from analyzing the prior solution. Similarly, successive prioritizations are generated by constructing and analyzing a solution, and then using the blame that results from that analysis to modify the previous prioritization.

A point in the solution space represents a potential solution to the problem, and a corresponding point in priority space, derived by analyzing the solution, is an attempt to capture information about the structure of the search space in the vicinity of the solution. As SWO constructs a new solution from scratch, the priorities can be thought of as providing information about pitfalls common to the current region of the solution space. If some elements of the solution have tended to be sources of difficulty over some number of iterations, increasing their priority makes it more likely that the constructor will handle those elements in a good way.

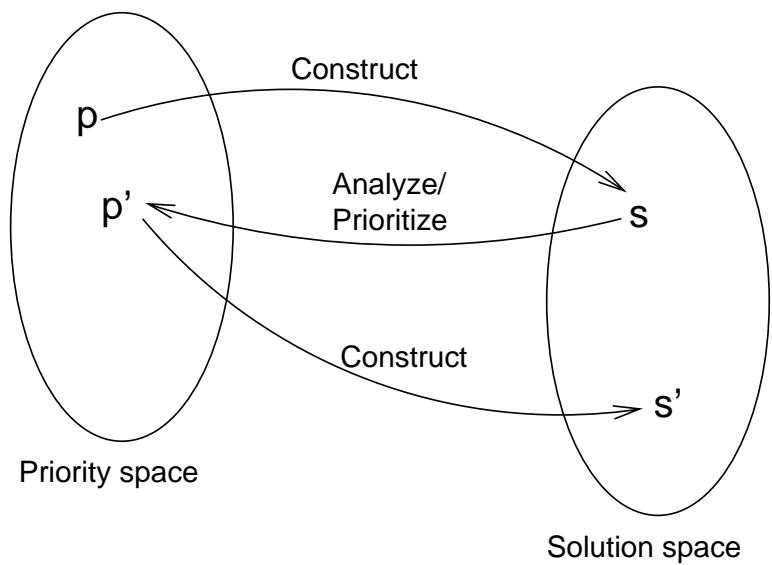

Figure 4: Coupled search spaces

One consequence of the coupled search spaces is that a small change in the sequence of elements generated by the prioritizer may correspond to a large change in the corresponding solution generated by the constructor, compared to the solution from the previous iteration. Moving an element forward in the sequence can significantly change its state in the resulting solution. In addition, any elements that now occur after it in the sequence must accommodate that element's state. For example, in the scheduling domain, moving a task earlier in the priority sequence may allow it to be placed on a different manufacturing line, thus possibly changing the mix of jobs that can run on that line, and on the line it was scheduled on in the previous iteration. One small change can have consequences for any element that follows it, with lower-priority tasks having to "fill in the gaps" that are left after higher-priority tasks have been scheduled.

The result is a large move that is "coherent" in the sense that it is similar to what we might expect from moving the higher priority task, then propagating the effects of that change by moving lower priority tasks as needed. This single move may correspond to a large number of moves for a search algorithm that only looks at local changes to the solution, and it may thus be difficult for such an algorithm to find.

The fact that SWO makes large moves in both search spaces is one obvious difference between SWO and traditional local search techniques, such as WSAT (Selman, Kautz, & Cohen, 1993). Another difference is that with SWO, moves are never selected based on their effect on the objective function. Instead, unlike hillclimbing techniques, each move is made in response to "trouble spots" found in the current solution. The resulting move may be uphill, but the move is always motivated by those trouble spots.

In priority space the only "local optima" are those in which all elements of a solution are assigned equal blame. SWO tends to avoid getting trapped in local optima, because analysis and prioritization will always (in practice) suggest changes in the sequence, thus changing the solution generated on the next iteration. This does not guarantee that SWO will not become trapped in a small cycle, however. In our implementations we have introduced small amounts of randomness in the basic cycle. We also restart SWO periodically with a new initial sequence.

Another aspect of local search is that typically each point in the solution space is associated with a single value, the objective function score for that solution. When we talk about hillclimbing, we generally refer to the "terrain" described by this objective function score, over the space of solutions. The process of analysis in SWO can be thought of as synthesizing a more complex description of that terrain, by breaking a solution down into its component elements and assigning a score to each. Prioritization then translates the analysis into a "strategy" that the constructor can use to generate the next solution.

Assigning scores to the individual elements of a solution allows SWO to take advantage of the fact that real problems often combine some elements that are difficult to get right, plus others that are easy. In the scheduling problems presented below, some tasks can be assigned to just a few production lines, while others allow for much more flexibility. Some have due dates close to their release time, while others have a lot of leeway. It is sometimes possible to identify "difficult" elements of a problem with static analysis, but interactions can be complex, and elements that are causing difficulty in one part of the search space may be no trouble at all in another. Rather than trying to identify elements that are globally difficult by analyzing the entire problem, SWO analyzes individual solutions in order to find elements that are *locally* difficult. Globally difficult elements tend to be identified over time, as they are difficult across large parts of the search space.

By assigning blame and adjusting priorities based on identified problems in actual solutions, SWO avoids dependence on complex, domain dependent heuristics. It is our belief that this independence is particularly important in complex domains where even the best heuristics will miss some key interactions and therefore inhibit the search from exploring good areas that the heuristic incorrectly labels as unpromising. SWO uses actual solutions to discover which areas of the search space are promising and which are not.

## 3. SWO for scheduling

This section describes an application of SWO to a fiber-optic production line scheduling problem, derived from data provided by Lucent Technologies. In this particular plant, a cable may be assembled on any one of 13 parallel production lines. For each cable type, only a subset of the production lines are compatible, and the time required to produce the cable will depend on which of the compatible lines is selected. Each cable also has a setup time, which depends on its own cable type and that of its predecessor. Setups between certain pairs of cable types are infeasible. Task preemption is not allowed, i.e. once a cable has started processing on a line, it finishes without interruption.

Each cable is assigned a release time and due date. Production cannot begin before the release time. The objective function includes a penalty for missing due dates, and a penalty for setup times.

## 3.1 Implementation

We describe the implementation in terms of the three main components of SWO:

**Constructor.** The constructor builds a schedule by adding tasks one at a time, in the order they occur in the priority sequence. A task is added by selecting a line and a position relative to the tasks already in that line. A task may be inserted between any two tasks already in the line or at the beginning or end of that line's schedule. Changes to the relative positions of the tasks already in the line are not considered. Each task in the line is then assigned to its earliest possible start time, subject to the ordering, i.e., a task starts at either its release time, or immediately after the previous task on that line, whichever is greater.

For each of the possible insertion points in the schedule, relative to the tasks already in each line, the constructor calculates the effect on the objective function, and the task is placed at the best-scoring location. Ties are broken randomly. After all tasks have been placed, the constructor applies SWO to the individual line schedules, attempting to improve the score for each line by reordering the cables that were assigned to it.

**Analyzer.** To assign blame to each task in the current schedule, the analyzer first calculates a lower bound on the minimum possible cost that each task could contribute to any schedule. For example, if a task has a release time that is later than its due date, then it will be late in every schedule, and the minimum possible cost already includes that penalty. Minimum possible setup costs are also included. For a given schedule, the blame assigned to each task is its "excess cost," the difference between its actual cost and its minimum possible cost. Excess lateness costs are assigned to tasks that are late, and excess setup costs are split between adjacent tasks.

**Prioritizer.** Once the blame has been assigned, the prioritizer modifies the previous sequence of tasks by moving tasks with non-zero blame factors forward in the sequence. Tasks are moved forward a distance that increases with the magnitude of the blame. To move from the back of the sequence to the front, a task must have a high blame factor over several iterations. We call this a "sticky sort."

Our current implementation has considerable room for improvement. The analysis and feedback currently being used are very simple, and the construction of schedules could take various heuristics into account, such as preferring to place a task in a line that has more "slack," all other things being equal.

## 3.2 Experimental results

We have six sets of test data, ranging in size from 40 to 297 tasks, all with 13 parallel production lines. The largest problem was the largest that the manufacturer required in practice. We compare the following solution methods:

| Data Set | Best Obj | SWO | | TABU | | IP | |
|---|---|---|---|---|---|---|---|
| | | Avg Obj | Avg Time | Obj | Time | Obj | Time |
| 40 | 1890 | 1890 | 48 | 1911 | 425 | 1934 | 20 |
| 50 | 3101 | 3156 | 57 | 3292 | 732 | 3221 | 175 |
| 60 | 2580 | 2584 | 87 | 2837 | 1325 | 2729 | 6144 |
| 70 | 2713 | 2727 | 124 | 2878 | 2046 | 2897 | 4950 |
| 148 | 8869 | 8927 | 431 | 10421 | 17260 | — | — |
| 297 | 17503 | 17696 | 1300 | — | — | — | — |

Table 1: Experimental results: scheduling

SWO Applies the SWO architecture to the problem, running for a fixed number of iterations and returning the best schedule it finds.

TABU Uses TABU search (Glover & Laguna, 1997), a local search algorithm in which moves that increase cost are permitted to avoid getting trapped at local optima. To avoid cycling, when an "uphill" move is made, it is not allowed to be immediately undone.

IP Applies an Integer Programming (IP) solver, using an encoding described in (Clements et al., 1997).

On the 297 task problem, SWO was far more effective than either TABU or IP. TABU, for example, failed to find a feasible schedule after running for over 24 hours. On the smallest problems, TABU and IP were able to find solutions, but SWO outperformed both by a substantial margin.

Table 1 presents results on each problem for SWO, TABU and IP. For SWO, ten trials were run and the results averaged. The TABU and IP implementations were deterministic, so only the results of a single run are shown. The second column of the table shows the best objective function value we have ever observed on each problem. The remaining columns show the objective function value and running times for SWO, TABU and IP. All but the IP experiments were run on a Sun Sparcstation 10 Model 50. The IP experiments were run on an IBM RS6000 Model 590 (a faster machine).

The best values observed have been the result of combining SWO with IP, as reported in (Clements et al., 1997). In that work, SWO generated solutions, running until it had produced a number of "good" schedules. An IP solver was then invoked to re-combine elements of those solutions into a better solution. Although the improvements achieved by the IP solver were relatively small, on the order of 1.5%, it achieved this improvement quickly, and SWO was unable to achieve the same degree of optimization even when given substantially more time. While noting that the hybrid approach can be more effective than SWO alone, and much more effective than IP alone, here we focus on the performance of the individual techniques.

We also note that our very first, fairly naive implementation of SWO for these scheduling problems already outperformed both TABU and IP. Moreover, our improved implementation, reported above, is still fairly simple, and is successful without relying on domain-dependent heuristics. We take this as evidence that the effectiveness of our approach is not due to cleverness in the construction, analysis and prioritization techniques, but due to the effectiveness of the swo cycle at identifying and responding to whatever elements of the problem happen to be causing difficulty in the local region of the search.

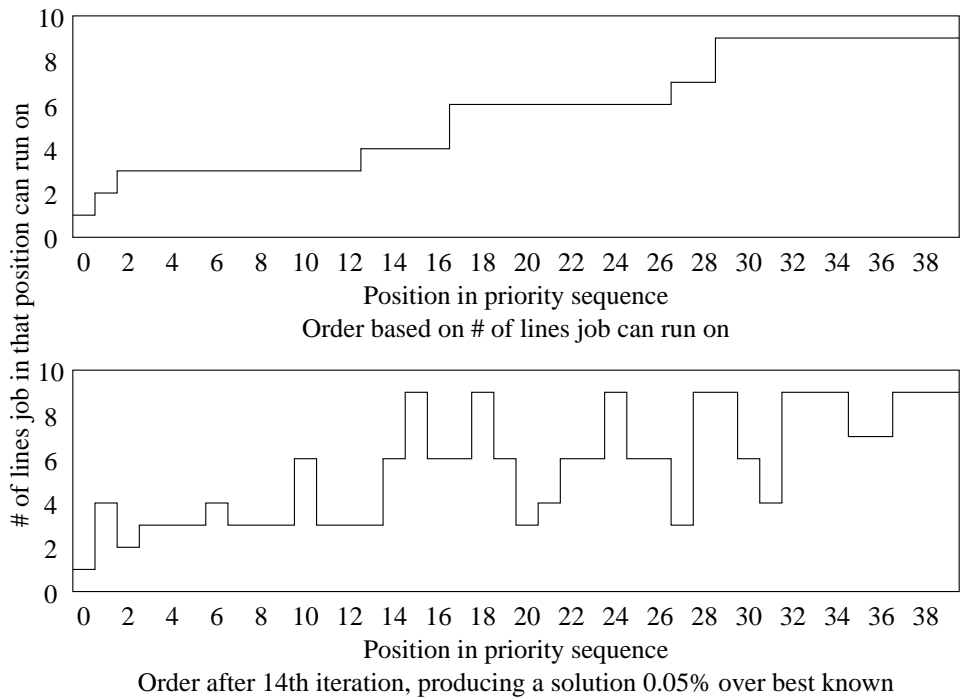

Figure 5: Comparison of heuristic priorities and priorities derived by swo

It is also instructive to compare the results of a good heuristic ordering, with the sequence derived by swo. A good heuristic for this scheduling domain (and the one that is used to initially populate the priority sequence) is to sort the tasks by the number of production lines on which a task could be feasibly assigned in an empty schedule. A task that can be scheduled on many lines is likely to be easier to schedule than one that is compatible with only a small number of lines, and should therefore be expected to need a lower priority. The top graph in Figure 5 shows the sequence of tasks, as determined by this heuristic. The lower graph illustrates the changes in priority of these tasks, after swo has run for fourteen iterations (enough to improve the solution derived from the sequence to within 0.05 percent of the best known solution).

As the figure illustrates, the heuristic is generally accurate, but swo has had to move some tasks that are compatible with most of the production lines to positions of relatively high priority, reflecting the fact that contrary to the heuristic, these tasks turned out to be relatively difficult to schedule well. Other tasks that are compatible with only a few production lines are actually easy to schedule well, and have moved to relatively low priorities.

| Iterations per Restart | Feasible | | < 18000 | | < 17700 | | Sample Size |
|---|---|---|---|---|---|---|---|
| | Success Rate | Mean Cost | Success Rate | Mean Cost | Success Rate | Mean Cost | |
| 10 | 0.8542 | 5.9 | 0.0504 | 195.3 | 0.0002 | 49994.5 | 10000 |
| 20 | 0.9722 | 6.0 | 0.2052 | 90.9 | 0.0006 | 33328.3 | 5000 |
| 30 | 0.9955 | 5.8 | 0.3812 | 67.5 | 0.0030 | 9895.5 | 3300 |
| 40 | 0.9996 | 5.8 | 0.5488 | 56.7 | 0.0060 | 6658.2 | 2500 |
| 50 | 0.9995 | 6.0 | 0.6330 | 57.0 | 0.0160 | 3112.7 | 2000 |
| 60 | 1.0000 | 5.7 | 0.7242 | 52.9 | 0.0188 | 3170.4 | 1650 |
| 70 | 1.0000 | 5.7 | 0.8079 | 50.2 | 0.0350 | 1973.5 | 1400 |
| 80 | 1.0000 | 6.2 | 0.8552 | 49.5 | 0.0296 | 2670.0 | 1250 |
| 90 | 1.0000 | 5.8 | 0.8827 | 48.9 | 0.0300 | 2965.3 | 1100 |
| 100 | 1.0000 | 5.9 | 0.8840 | 52.4 | 0.0400 | 2452.3 | 1000 |
| 200 | 1.0000 | 6.0 | 0.9680 | 53.0 | 0.0600 | 3204.3 | 500 |
| 300 | 1.0000 | 5.3 | 0.9967 | 50.1 | 0.0567 | 5090.8 | 300 |
| 400 | 1.0000 | 5.8 | 1.0000 | 52.9 | 0.0720 | 5320.2 | 250 |
| 500 | 1.0000 | 5.8 | 1.0000 | 52.8 | 0.1000 | 4692.6 | 200 |
| 600 | 1.0000 | 5.8 | 1.0000 | 57.2 | 0.0867 | 6590.8 | 150 |
| 700 | 1.0000 | 6.1 | 1.0000 | 42.4 | 0.1200 | 5472.4 | 100 |
| 800 | 1.0000 | 5.6 | 1.0000 | 53.0 | 0.1200 | 6210.3 | 100 |
| 900 | 1.0000 | 5.3 | 1.0000 | 45.8 | 0.1700 | 4691.6 | 100 |
| 1000 | 1.0000 | 6.0 | 1.0000 | 45.4 | 0.1800 | 4838.1 | 100 |

Table 2: Experimental results: restarts in the scheduling domain

## 3.3 Restarts

The SWO solver used to produce the results reported in Table 1 restarted the priority queue every $n/2$ iterations, where $n$ is the number of jobs in the problem. The same noisy heuristic that was used to initially populate the priority queue was also used to restart it. This restart cutoff was picked in a rather ad hoc manner. A more careful analysis of different restart cutoff values might lead to producing better solutions faster, and to some additional insight on the workings of SWO.

Restarts are often used in non-systematic search to avoid getting trapped in local optima or in cycles. (See Parkes and Walser, 1996, for an empirical study of WSAT and further references.) Restarts have also been used in systematic search to escape exponentially large regions of the search space that do not contain a solution (Gomes, Selman, & Kautz, 1998).

Local optima pose little threat to SWO, since it is not directly driven by uphill/downhill considerations. SWO, through its use of large coherent moves, also tends to escape unpromising parts of the search space quickly. However, SWO is open to getting trapped in a cycle, and restarts are used as a means to escape them.

For these scheduling problems, SWO is unlikely to get into a tight cycle where priority queues and solutions repeat exactly. This is due to the presence of random tie breaking in several places, and to the presence of noise in the prioritizer. However, it is our belief that SWO can get trapped in a cycle where similar priority queues and solutions repeat.

We ran a series of experiments with the 297 task problem to determine the impact of various restart cutoffs. The results are summarized in Table 2. Restart cutoffs ranged from after every 10 iterations to after every 1000 iterations. The success rate and mean cost are shown for each value for each of three different solution qualities. The success rate indicates the probability that a solution of at least the given quality was found in a given pass. The mean cost is the average number of total iterations to get a solution of that quality.

For the feasible and 18000 solution thresholds, SWO reaches a 100 percent success rate well before reaching the maximum restart cutoff of 1000 used in these experiments. In some sense, it is easy for SWO to produce solutions that are at least of these qualities. The results for these 2 thresholds indicate that when it is easy for SWO to solve the problem, any cutoff greater than the average number of uninterrupted iterations it takes to produce a solution can be used to solve the problem at minimum cost. For such "easy" problems, it appears that too small a restart cutoff can hurt, but that too big a cutoff will not.

The numbers for the 17700 solution quality threshold, tell a different story. The success rate is still climbing when the experiment ends, and the mean cost has actually risen above its minimum. For this solution quality, the restart cutoff that minimizes mean cost falls around the range of 70 to 100. Mean costs rise steeply for restart cutoffs below this range, and slowly for cutoffs larger than that. This is an example of a hard problem for SWO, and it shows that some care needs to be taken when choosing a restart strategy for such problems. Additional research is needed to determine how to set the restart cutoff automatically for arbitrary problems.

This data indicates that SWO does benefit from restarts, up to a point. With the 17700 threshold, for restart cutoffs up to 100, each increase in the cutoff in general led to a superlinear increase in the success rate. (This is also another indicator that SWO is learning from iteration to iteration.) Above 100 iterations per restart, the success rate initially climbs sublinearly and then appears to level out. It is an open question what this tells us about the search space.

## 4. SWO for graph coloring

We have also applied SWO to a very different domain, graph coloring. Here the objective is to color the nodes of a graph such that no two adjoining nodes have the same color, minimizing the number of colors.

### 4.1 Implementation

The priority sequence for graph coloring consists of an ordered list of nodes. The solver is always trying to produce a coloring that uses colors from the *target set*, which has one less color than was used to color the best solution so far. Again, we describe the implementation in terms of the three main components of SWO:

**Constructor.** The constructor assigns colors to nodes in priority sequence order. If a node's color in the previous solution is still available (i.e. no adjacent node is using it yet), and is in the target set, then that color is assigned. If that fails, it tries to assign a color in the current target set, picking the color that is least constraining on adjacent uncolored nodes, i.e. the color that reduces the adjacent nodes' remaining color choices the least. If none of the target colors are available, the constructor tries to "grab" a color in the target set from its neighbors. A color can only be grabbed if all neighbor nodes with that color have at least one other choice within the target set. If multiple colors can be grabbed, then the least constraining one is picked. If no color in the target set can be grabbed then a color outside the target set is assigned.

Nodes that are early in the priority sequence are more likely to have a wide range of colors to pick from. Nodes that come later may grab colors from earlier nodes, but only if the earlier nodes have other color options within the target set.

**Analyzer.** Blame is assigned to each node whose assigned color is outside the target set, with the amount of blame increasing for each additional color that must be added to the target set. We ran experiments with several different variations of color-based analysis. All of them performed reasonably.

**Prioritizer.** The prioritizer modifies the previous sequence of nodes by moving nodes with blame forward in the sequence according to how much blame each received. This is done the same way it is done for the scheduling problems. The initial sequence is a list of nodes sorted in decreasing degree order, with some noise added to slightly shuffle the sort.

## 4.2 Experimental results

We applied SWO to a standard set of graph coloring problems, including random graphs and application graphs that model register allocation and class scheduling problems. These were collected for the Second DIMACS Implementation Challenge (Johnson & Trick, 1996), which includes results for several algorithms on these problems (Culberson & Luo, 1993; Glover, Parker, & Ryan, 1993; Lewandowski & Condon, 1993; Morgenstern, 1993). Problems range from 125 nodes with 209 edges to 4000 nodes with 4,000,268 edges.

Glover et al. (1993) is the only paper that is based on a general search technique, TABU with branch and bound, rather than a graph coloring specific algorithm. This approach had the worst reported average results in the group. Morgenstern (1993) used a distributed IMPASSE algorithm and had the best overall colorings, but also required that the target number of colors, as well as several other problem specific parameters be passed to the solver. Lewandowski & Condon (1993) also found good solutions for this problem set. Their approach used a hybrid of parallel IMPASSE and systematic search on a 32 processor CM-5. Culberson & Luo (1993) used an Iterated Greedy (IG) algorithm that bears some similarity to SWO. IG is the simplest algorithm in the group. Its solution quality falls between the IMPASSE algorithms and TABU but solves the entire set in 1 to 2 percent of the time taken by the other methods. Both IG and IMPASSE are discussed further under related work.

Table 3 compares SWO with the results for IG (Culberson & Luo, 1993), distributed IMPASSE (Morgenstern, 1993), parallel IMPASSE (Lewandowski & Condon, 1993), and TABU (Glover et al., 1993). For each, one column shows the number of colors required for each problem, and the run time (in CPU seconds). Bold face indicates that the number of colors is within 0.5 of the best result in the table.

We used a Pentium Pro 333MHz workstation running Linux for the SWO graph coloring experiments. The times shown for the other four algorithms are based on those reported in (Johnson & Trick, 1996). The results for IG, IMPASSE and TABU are normalized to our times using the DIMACS benchmarking program *dfmax*, provided for this purpose. Therefore, timing comparisons are only approximate. Our machine ran the *dfmax r500.5* benchmark in 86.0 seconds; the times reported for the machines used on the other algorithms were 86.9 seconds for the TABU experiments, 192.6 seconds for IG, 189.3 seconds for IMPASSE, and 2993.6 seconds for parallel IMPASSE. Because the *dfmax* benchmark runs on a single processor, it is unsuitable for normalizing the times for parallel IMPASSE. We report their unnormalized times.

| Problem | SWO | | IG | | Dist. IMPASSE | | Par. IMPASSE | | TABU | |
|---|---|---|---|---|---|---|---|---|---|---|
| | colors | time | colors | time | colors | time | colors | time | colors | time |
| DSJC125.5 | 18.3 | 1.6 | 18.9 | 2.5 | **17.0** | 6.3 | **17.0** | 4043.6 | 20.0 | 153.3 |
| DSJC250.5 | 31.9 | 8.3 | 32.8 | 6.9 | **28.0** | 268.5 | 29.2 | 4358.1 | 35.0 | 3442.2 |
| DSJC500.5 | 56.3 | 40.9 | 58.6 | 18.2 | **49.0** | 8109.1 | 53.0 | 4783.9 | 65.0 | 3442.2 |
| DSJC1000.5 | 101.5 | 208.6 | 104.2 | 67.6 | **89.0** | 41488.7 | 100.0 | 5333.8 | 117.0 | 3442.2 |
| C2000.5 | 185.7 | 1046.2 | 190.0 | 272.4 | **165.0** | 14097.9 | — | — | — | — |
| C4000.5 | **341.6** | 4950.8 | 346.9 | 1054.1 | — | — | — | — | — | — |
| R125.1 | **5.0** | 0.2 | **5.0** | 2.0 | **5.0** | 0.2 | **5.0** | 64.6 | **5.0** | 0.4 |
| R125.1c | **46.0** | 5.1 | **46.0** | 1.1 | **46.0** | 0.2 | **46.0** | 85.0 | **46.0** | 0.9 |
| R125.5 | **36.0** | 2.8 | 36.9 | 1.9 | **36.0** | 0.2 | 37.0 | 33.0 | **36.0** | 0.7 |
| R250.1 | **8.0** | 0.5 | **8.0** | 7.0 | **8.0** | 0.2 | **8.0** | 22.0 | **8.0** | 0.2 |
| R250.1c | **64.0** | 30.6 | **64.0** | 4.6 | **64.0** | 0.5 | **64.0** | 278.2 | 65.0 | 46.4 |
| R250.5 | **65.0** | 14.7 | 68.4 | 8.3 | **65.0** | 82.2 | 66.0 | 39.9 | 66.0 | 59.0 |
| DSJR500.1 | **12.0** | 2.0 | **12.0** | 21.1 | **12.0** | 0.2 | **12.0** | 26.6 | **12.0** | 0.5 |
| DSJR500.1c | **85.2** | 96.9 | **85.0** | 14.6 | **85.0** | 59.1 | **85.2** | 5767.7 | 87.0 | 3442.2 |
| DSJR500.5 | 124.1 | 68.7 | 129.6 | 26.1 | **123.0** | 175.3 | 128.0 | 90.5 | 126.0 | 395.1 |
| R1000.1 | **20.0** | 8.0 | 20.6 | 87.2 | **20.0** | 8.2 | **20.0** | 49.9 | **20.0** | 1.7 |
| R1000.1c | 101.7 | 433.2 | 99.8 | 49.1 | **98.0** | 563.3 | 102.6 | 3940.0 | 105.0 | 3442.2 |
| R1000.5 | **238.9** | 574.5 | 253.2 | 102.9 | 241.0 | 944.0 | 245.6 | 215.9 | 248.0 | 3442.2 |
| flat300_20_0 | 25.3 | 16.4 | **20.2** | 3.8 | **20.0** | 0.2 | **20.0** | 274.3 | 39.0 | 3442.2 |
| flat300_26_0 | 35.8 | 12.0 | 37.1 | 7.7 | **26.0** | 10.0 | 32.4 | 6637.1 | 41.0 | 3442.2 |
| flat300_28_0 | 35.7 | 11.9 | 37.0 | 9.6 | **31.0** | 1914.2 | 33.0 | 1913.5 | 41.0 | 3442.2 |
| flat1000_50_0 | 100.0 | 203.9 | 65.6 | 146.3 | **50.0** | 0.2 | 97.0 | 7792.7 | — | — |
| flat1000_60_0 | 100.7 | 198.0 | 102.5 | 87.3 | **60.0** | 0.2 | 97.8 | 6288.4 | — | — |
| flat1000_76_0 | 100.6 | 208.4 | 103.6 | 79.6 | **89.0** | 11034.0 | 99.0 | 6497.9 | — | — |
| latin_sqr_10 | 111.5 | 369.2 | 106.7 | 59.7 | **98.0** | 5098.0 | 109.2 | 6520.1 | 130.0 | 3442.2 |
| le450_15a | **15.0** | 5.5 | 17.9 | 17.0 | **15.0** | 0.2 | **15.0** | 162.6 | 16.0 | 17.8 |
| le450_15b | **15.0** | 6.1 | 17.9 | 16.2 | **15.0** | 0.2 | **15.0** | 178.4 | **15.0** | 28.4 |
| le450_15c | 21.1 | 8.0 | 25.6 | 14.5 | **15.0** | 57.2 | 16.6 | 2229.6 | 23.0 | 3442.2 |
| le450_15d | 21.2 | 7.8 | 25.8 | 13.5 | **15.0** | 36.3 | 16.8 | 2859.6 | 23.0 | 3442.2 |
| mulsol.i.1 | **49.0** | 5.9 | **49.0** | 4.2 | **49.0** | 0.2 | **49.0** | 27.2 | **49.0** | 0.3 |
| school1 | **14.0** | 8.4 | **14.0** | 10.5 | **14.0** | 0.2 | **14.0** | 46.3 | 29.0 | 90.7 |
| school1_nsh | **14.0** | 7.2 | **14.1** | 8.9 | **14.0** | 0.2 | **14.0** | 66.4 | 26.0 | 31.2 |

Table 3: Experimental results: graph coloring problems

A variety of termination conditions were used. SWO terminated after 1000 iterations. IG terminated after 1000 iterations without improvement. Distributed IMPASSE used a wide variety of different termination conditions to solve the different problems. The only common element across problems was that distributed IMPASSE stopped when the target number of colors, provided as an input parameter, was reached. The times reported for parallel IMPASSE are the times it took to find the best solution that was found, not the time it took the algorithm to terminate, which was always 3 hours. TABU ran until the algorithm determined that it could make no further progress, or an hour had passed, whichever came first.

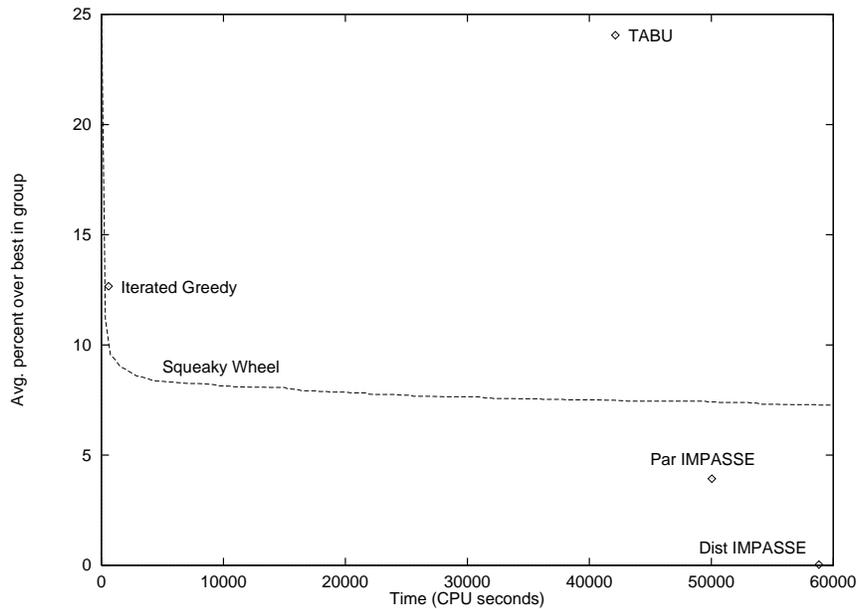

Figure 6: Experimental results: quality of solution *vs.* time

The TABU numbers are for a single run on each problem. The numbers for the other algorithms are averages for 4 runs (parallel IMPASSE), 5 runs (distributed IMPASSE, parallel IMPASSE) or 10 runs (SWO, IG, distributed IMPASSE) on each problem.

Figure 6 summarizes the performance of each technique on the set of 27 problems that all of the algorithms solved. For each solver the graph indicates the average solution quality and the average amount of time needed to solve the set. The ideal location on the graph is the origin, producing high quality solutions in very little time. The points shown for the other techniques are the points reported in each of the papers. The curve shown for SWO shows how it performs when given varying amounts of time to solve the set. As the graph shows, SWO clearly outperforms TABU, the only other general purpose technique, both in terms of quality and speed. SWO also outperforms IG, a graph coloring specific algorithm, both in terms of quality and speed. The IMPASSE solvers clearly produce the best solutions in the group. However, IMPASSE is a domain specific method, and both solvers represent much more programming effort. The SWO solver uses a general purpose search technique and was implemented in less than a month by a single programmer.

## 4.3 Alternate configurations of SWO

We note that, as with the scheduling work, our first, naive implementation of SWO for graph coloring produced respectable results. Even without color reuse, color grabbing, or the least constraining heuristic (the first free color found was picked), SWO matched IG on 6 problems and beat it on 10. However, on half of the remaining problems IG did better by 10 or more colors.

To explore the sensitivity of SWO to such implementation details we tried the following approaches in the constructor and prioritizer, and ran SWO using all combinations:

**Construction:** With or without color grabbing

**Analysis:** Either blame all nodes that receive a color outside the target set, or only the first node (in the priority sequence) that causes a new color outside the target set to be introduced. If color grabbing is used, the determination of blame is based on the final color assigned to the node.

The difference in solution quality from the worst combination to the best combination was less than 15 percent. Even when the alternative of using a standard sort instead of the "sticky" sort (a fairly fundamental change) was added to the mix, the spread between worst and best was still under 20 percent.

## 5. Related work

The importance of prioritization in greedy algorithms is not a new idea. The "First Fit" algorithm for bin packing, for example, relies on placing items into bins in decreasing order of size (Garey & Johnson, 1979). Another example is GRASP (Greedy Randomized Adaptive Search Procedure) (Feo & Resende, 1995). GRASP differs from our approach in several ways. First, the prioritization and construction aspects are more closely coupled in GRASP. After each element is added to the solution being constructed, the remaining elements are re-evaluated by some heuristic. Thus the order in which elements are added to the solution may depend on previous decisions. Second, the order in which elements are selected in each trial is determined only by the heuristic (and randomization), so the trials are independent. There is no learning from iteration to iteration in GRASP.

Doubleback Optimization (DBO) (Crawford, 1996) was to some extent the inspiration for both SWO and another similar algorithm, Abstract Local Search (ALS) (Crawford, Dalal, & Walser, 1998). In designing SWO, we began by looking at DBO, because it had been extremely successful in solving a standard type of scheduling problem. However, DBO is only useful when the objective is to minimize makespan, and is also limited in the types of constraints it can handle. Because of these limitations, we began thinking about the principles behind DBO, looking for an effective generalization of that approach. DBO can, in fact, be viewed as an instance of SWO. DBO begins by performing a "right shift" on a schedule, shifting all tasks as far to the right as they can go, up to some boundary. In the resulting right-shifted schedule, the left-most tasks are, to some extent, those tasks that are most critical. This corresponds to analysis in SWO. Tasks are then removed from the right-shifted schedule, taking left-most tasks first. This ordering corresponds to the prioritization in SWO. As each task is removed, it is placed in a new schedule at the earliest possible start time, i.e., greedy construction.

Like SWO, ALS was the result of an attempt to generalize DBO. ALS views priority space (to use the terminology from SWO) as a space of "abstract schedules," and performs a local search in that space. Unlike SWO, if a prioritization is modified, and the corresponding move in solution space is downhill (away from optimal), then the modified prioritization is discarded, and the old prioritization is restored. As is usual with local search, ALS also sometimes makes random moves, in order to escape local minima.

ALS, and also List Scheduling (Pinson, Prins, & Rullier, 1994), are scheduling algorithms that deal with domains that include precedence constraints on tasks. Both accommodate precedence constraints by constructing schedules left-to-right temporally. A task cannot be placed in the schedule until all of its predecessors have been placed. In order for the analysis, prioritization and construction to be appropriately coupled, it is not sufficient to simply increase the priority of a task that is late, because the constructor may not be able to place that task until after a lot of other decisions have been made. Consequently, some amount of blame must be propagated to the task's predecessors.

The commercial scheduler OPTIFLEX (Syswerda, 1994) uses a genetic algorithm approach to modify a sequence of tasks, and a constraint-based schedule constructor that generates schedules from those sequences. OPTIFLEX can also be viewed as an instance of SWO, with a genetic algorithm replacing analysis. In effect, the "analysis" instead emerges from the relative fitness of the members of the population.

Two graph coloring algorithms also bear some similarity to SWO. Impasse Class Coloration Neighborhood Search (IMPASSE) (Morgenstern, 1993; Lewandowski & Condon, 1993), like SWO, maintains a target set of colors and produces only feasible colorings. Given a coloring, IMPASSE places any nodes that are colored outside of the target set into an impasse set. On each iteration a node is selected from the impasse set, using a noisy degree-based heuristic, and assigned a random color from the target set. Any neighbor nodes that are now in conflict are moved to the impasse set.

Iterated Greedy (IG) (Culberson & Luo, 1993), like SWO, uses a sequence of nodes to create a new coloring on each iteration, and then uses that coloring to produce a new sequence for the next iteration. The method used to generate each new sequence differs from SWO. The key observation behind IG is that if all nodes with the same color in the current solution are grouped together in the next sequence (i.e. adjacent to each other in the sequence), then the next solution will be no worse than the current solution. IG achieves improvement by manipulating the order in which the groups occur in the new sequence, using several heuristics including random based on color, descending based on color, and ascending based on the cardinality of each group. IG learns groupings of nodes as it runs, but it does not learn about about the difficulty of any nodes. A node's place in the sequence indicates nothing about its expected or detected difficulty.

## 6. Analysis and future work

This section summarizes several areas of future research suggested by the results reported in the previous sections.

### 6.1 Scaling

While SWO uses fast, greedy algorithms for constructing solutions, and we have demonstrated its effectiveness on problems of realistic size, the greatest threat to the scalability of SWO is that it constructs a new solution from scratch on each iteration. A partial solution to this problem is seen in the use of a "history" mechanism for the graph coloring problems.

Using the same color for a node as in the previous solution means that in many cases we do not need to check any of the other possible colors. This significantly speeds up the construction.

A more fundamental solution to this problem would be to develop an incremental version of SWO. The selective reuse of colors in the graph coloring solver is a small step in this direction. This allows the constructor to avoid spending time evaluating other alternatives when the previous choice still works. More generally, it may be possible to look at the changes made to a prioritization, and modify the corresponding solution in a way that generates the same solution that would be constructed from scratch based on the new prioritization. It seems feasible that this could be done for some domains, at least for small changes to the prioritization, because there may be large portions of a solution that are unaffected.

A more interesting possibility is based on the view of SWO as performing local search plus a certain kind of propagation. A small change in priorities may correspond to a large change in the solution. For example, increasing the priority of one task in a scheduling problem may change its position in the schedule, and, as a consequence, some lower priority tasks may have to be shuffled around to accommodate that change. This is similar to what we might expect from moving the higher priority task, then propagating the effects of that change by moving lower priority tasks as well. This single move may correspond to a large number of moves in a search algorithm that only looks at local changes to the schedule, and may thus be difficult for such an algorithm to find.

Based on this view, we are investigating an algorithm we call "Priority-Limited Propagation" (PLP). With PLP, local changes are made to the solution, and then propagation is allowed to occur, subject to the current prioritization. Propagation is only allowed to occur in the direction of lower-priority elements. In effect, a small change is made, and then the consequences of that change are allowed to "ripple" through the plan. Because propagation can only occur in directions of decreasing priority, these ripples of propagation decrease in magnitude until no more propagation is possible. A new prioritization is then generated by analyzing the resulting solution. (It should be possible to do this analysis incrementally, as well.) The resulting approach is not identical to SWO, but has many of its interesting characteristics.

## 6.2 Coordination of modules

For SWO to be effective, it is obvious that analysis, prioritization and construction must all work together to improve the quality of solutions. We have already discussed the complications that can arise when constraints are placed on the order in which the constructor can make decisions, as is the case for List Scheduling and ALS, where construction is done strictly left-to-right. Without more complex analysis, the search spaces can effectively become uncoupled, so that changes in priority don't cause the constructor to fix problems discovered by analysis.

Another way the search can become uncoupled is related to the notion of "excess cost," discussed for the scheduling implementation. The calculation of excess cost in the analyzer turned out to be a key idea for improving the performance of SWO. However, problems sometimes have tasks that must be handled badly in order to achieve a good overall solu-tion. One of the scheduling problems described previously has two such "sacrificial" tasks. Whenever a good solution is found, the analyzer assigns high blame to these sacrificial tasks, and the constructor handles them well on the next iteration. This means that the resulting solution is of poor overall quality, and it is not until other flaws cause other tasks to move ahead of the sacrificial tasks in the priority sequence that SWO can again, briefly, explore the space of good solutions. In such cases, to some extent the analysis is actually hurting the ability of SWO to converge on good solutions.

Ideally, we would like to generalize the notion of excess cost to recognize sacrificial tasks, and allow those tasks to be handled badly without receiving proportionate blame. For problems in which a task must be sacrificed in *all* solutions, it may be possible to use a learning mechanism to accomplish this.

However, the notion of a sacrificial task can be more subtle than this. Suppose for example that we are scheduling the construction of two airplanes, *P1* and *P2*, and that each has a key task, *T1* and *T2*, respectively, requiring all of some shared resource, *R*. Because of the resource conflict, we must either give *R* to *T1* early in the schedule, starting construction on plane *P1* before *P2*, or we must give *R* to *T2* early in the schedule, with the opposite result. Whichever of the two tasks is started early will finish on time, but the other will be late.

Suppose we construct a schedule in which *T1* goes first, and *T2* is late, thus receiving a heavy blame factor. SWO increases the priority on *T2*, and as a consequence, *T2* goes first in the subsequent schedule. But then *T1* is late, and on the next iteration it will again go first. We could alternate in this manner forever, and the result would be that SWO would fail to explore either option very effectively, because it would be jumping back and forth between the option of building plane *P1* first, and the option of building plane *P2* first, without remaining in one region of the search space long enough to refine a solution.

The difficulty is that neither *T1* nor *T2* can be identified as a sacrificial task. Assuming the two planes are not identical, we cannot simply argue from symmetry that we should just pick one of the two tasks to be sacrificed. If, however, we could identify a sacrificial task by the *role* it plays in a solution, we could achieve what we need. Here, the task to be sacrificed must be the one that belongs to whichever plane is started later. If the analyzer could reduce the blame assigned to that task in a schedule, whichever task it happens to be, it would allow SWO to explore that region of the search much more effectively.

This problem of interchangeable roles would arise even more clearly with the introduction of conditional elements in a solution. Suppose, for example, we have a scheduling problem in which the constructor may choose to include or not include task instances of some type, adding however many instances are needed to satisfy a resource requirement. If those tasks are all instances of the same task type, then they are interchangeable, and penalizing one may simply cause a shuffling of those instances that does not really address the problem. Moreover, with conditional tasks, it is not clear how the analyzer should assign blame when the set of task instances in the current schedule may be very different from the set of task instances in successor schedules.

To address these concerns, the notion of prioritization could be generalized to apply to additional aspects of a problem. In scheduling this might mean not just prioritizing tasks, but also resources over various time intervals. We also propose that the these prioritizations be limited to the "fixed" elements of a problem. In scheduling problems, for example, these may be the non-conditional tasks, resources, etc. (In our example domains, all of the elements are fixed in this sense, so this was not an issue.)

One intuition behind this proposal is that these are the elements that will tend to define roles. In the earlier example with tasks *T1* and *T2*, corresponding to the two planes being built, the critical element is not either task *per se*, but actually resource *R*, early in the schedule. If this phase of resource *R* receives a high priority, and the later phase of resource *R* receives a lower priority, then whichever of the two tasks occurs later will be recognized as less critical. While this does not exactly capture the notion of "role" that we would like, it comes a lot closer than the current approach. In addition, assigning priorities to the fixed elements of a problem has the advantage of being applicable to problems with conditional tasks. Research is currently under way to explore this approach.

### 6.3 swo **and local search**

Although the ability to make large, coherent moves is a strength of the approach, it is also a weakness. swo is poor at making small "tuning" moves in the solution space, but the coupled-search view of swo suggests an obvious remedy. swo could be combined with local search in the solution space, to look for improvements in the vicinity of good solutions. Similarly, making small changes to a prioritization would generally result in smaller moves in the solution space than result from going through the full analysis and re-prioritization cycle.

Yet another alternative is genetic algorithm techniques for "crossover" and other types of mutation to a pool of nodes, as is done in OPTIFLEX. Many hybrid approaches are possible, and we believe that the coupled-search view of swo helps to identify some interesting strategies for combining moves of various sizes and kinds, in both search spaces, adapting dynamically to relative solution qualities.

## 7. Conclusions

Our experience has been that it is fairly straightforward to implement swo in a new domain, because there are usually fairly obvious ways to construct greedy solutions, and to analyze a solution to assign "blame" to some of the elements. Naive implementations of swo tend to perform reasonably well.

We have found the view of swo as performing a "coupled search" over two different search spaces to be very informative. It has been helpful to characterize the kinds of moves that swo makes in each of the search spaces, and the effect this has on avoiding local optima, etc. We hope that by continuing to gain a deeper understanding of what makes swo work we will be able to say more about the effective design of swo algorithms.

As the number of directions for future research suggests, we have only begun to scratch the surface of "Squeaky Wheel" Optimization.

### Acknowledgements

The authors wish to thank Robert Stubbs of Lucent Technologies for providing the data used for the scheduling experiments. The authors also wish to thank George L. Nemhauser,

Markus E. Puttlitz and Martin W. P. Savelsbergh with whom we collaborated on using SWO in a hybrid AI/OR approach. Many useful discussions came out of that collaboration, and without them we would not have had access to the Lucent problems. Markus also wrote the framework for the scheduling experiments and the TABU and IP implementations.

The authors also thank the members of CIRL, and James Crawford at i2 Technologies, for their helpful comments and suggestions. We would like to thank Andrew Parkes in particular for suggestions and insights in the graph coloring domain.

This effort was sponsored by the Air Force Office of Scientific Research, Air Force Materiel Command, USAF, under grant number F49620-96-1-0335; by the Defense Advanced Research Projects Agency (DARPA) and Rome Laboratory, Air Force Materiel Command, USAF, under agreements F30602-95-1-0023 and F30602-97-1-0294; and by the National Science Foundation under grant number CDA-9625755.

The U.S. Government is authorized to reproduce and distribute reprints for Governmental purposes notwithstanding any copyright annotation thereon. The views and conclusions contained herein are those of the authors and should not be interpreted as necessarily representing the official policies or endorsements, either expressed or implied, of the Defense Advanced Research Projects Agency, Rome Laboratory, the Air Force Office of Scientific Research, the National Science Foundation, or the U.S. Government.

Most of the work reported in this paper was done while both authors were at CIRL.